\documentclass{article}

\PassOptionsToPackage{numbers, compress}{natbib}

\usepackage{amsmath}
\usepackage{graphicx}

\usepackage[final]{neurips_2021}

\usepackage[utf8]{inputenc} 
\usepackage[T1]{fontenc}    
\usepackage{hyperref}       
\usepackage{url}            
\usepackage{booktabs}       
\usepackage{amsfonts}       
\usepackage{nicefrac}       
\usepackage{microtype}      
\usepackage{xcolor}         
\usepackage{wrapfig}
\pdfoutput=1
\title{Learning from Mistakes: Using Mis-predictions as Harm Alerts in Language Pre-Training}

\author{%
  Chen Xing, Wenhao Liu and Caiming Xiong \\
  Salesforce Research, Palo Alto, USA \\
{\tt\small{\{cxing,wenhao.liu,cxiong\}@salesforce.com}}\\
}

\begin{document}

\maketitle

\vskip 0.3in




\begin{abstract}

Fitting complex patterns in the training data, such as reasoning and commonsense, is a key challenge for language pre-training. 
According to recent studies and our empirical observations, one possible reason is that some easy-to-fit patterns in the training data, such as frequently co-occurring word combinations, dominate and harm pre-training, making it hard for the model to fit more complex information. 
We argue that mis-predictions can help locate such dominating patterns that harm language understanding. 
When a mis-prediction occurs, there should be frequently co-occurring patterns with the mis-predicted word fitted by the model that lead to the mis-prediction. 
If we can add regularization to train the model to rely less on such dominating patterns when a mis-prediction occurs and focus more on the rest more subtle patterns, more information can be efficiently fitted at pre-training. 
Following this motivation, we propose a new language pre-training method, Mis-Predictions as Harm Alerts (MPA). 
In MPA,  when a mis-prediction occurs during pre-training, 
we use its co-occurrence information to guide several heads of the self-attention modules. 
Some self-attention heads in the Transformer modules are optimized to assign lower attention weights to the words in the input sentence that frequently co-occur with the mis-prediction while assigning higher weights to the other words.
By doing so, the Transformer model is trained to rely less on the dominating frequently co-occurring patterns with mis-predictions while focus more on the rest more complex information when mis-predictions occur. Our experiments show that MPA  expedites the pre-training of BERT and ELECTRA and improves their performances on downstream tasks.
\end{abstract}

\section{Introduction}
\label{section:introduction}

Transformer models~\citep{vaswani2017attention} pre-trained by language modeling tasks are very powerful on a wide variety of NLP tasks~\citep{devlin2018bert, liu2019roberta}. 
However, the convergence efficiency of language pre-training methods hasn't been improved much~\citep{li2020train, kaplan2020scaling}. 
Given finite amount of time and computing resources, language pre-training methods can be very under-trained~\citep{liu2019roberta}. Therefore unlike most of classic deep learning tasks in which overfitting is a concern, the  challenge of language pre-training is actually underfitting the training data~\citep{liu2019roberta,carlini2020extracting,li2020train,shoeybi2019training}.
What makes data fitting so hard for language pre-training? Recent research shed light on this issue by interpreting the attention weights of pre-trained Transformer models\citep{clark2019does, rogers2020primer} or running probing tasks on pre-trained models with different situations\citep{liu2019linguistic,zhang2020you, talmor2019olmpics, tenney2019you, liu2021probing}. 
One common observation from such work is that pre-trained model's attention modules can be good at capturing linguistic information such as syntactic patterns, while having a hard time at fitting more complex patterns such as human commonsense and reasoning. This is probably due to that models pre-trained with language modeling tasks rely too much on word co-occurrence information \citep{tenney2019you,talmor2019olmpics}.
We observed similar phenomenon by manually examining the word predictions of Masked Language Model (MLM) at masked positions during pre-training. 
We found that even at the early stage of pre-training, the pre-trained model can surprisingly produce very fluent sentences. However, the model can have mis-predictions that semantically or logically contradict with the context through the entire pre-training. One example is,

\begin{center}
 \textbf{\texttt{He went back to his \underline{bedroom} to continue his draft design of a new bed.}}
\end{center}

In the masked position of the sentence above, the ground-truth word is ``study'', which is mis-predicted as ``bedroom''. It is probably because the frequently co-occurring pattern between ''bedroom'' and ''bed'' that is easy to fit for the model, dominates pre-training and outruns the hard-to-fit semantics in the context.
Such easy-to-fit patterns can therefore prevent the model from fitting the more sophisticated patterns, such as reasoning and rare facts, and harm the model's performance on downstream tasks.

Fortunately, we believe that mis-predictions can help locate such dominating patterns the model has fitted that harm language understanding. When a mis-prediction occurs, there are likely to be some dominating patterns related to the mis-prediction in the context fitted by the model that cause this mis-prediction, for example, the frequently co-occurring word ''bed'' with the mis-prediction ''bedroom'' in the highlighted example. If we can add regularization to train the model to rely less on these dominating patterns such as word co-occurrences when a mis-prediction occurs, thus focusing more on the rest more subtle patterns, more information can be efficiently fitted at pre-training. Following this motivation, we propose a new language pre-training method, Using Mis-predictions as Harm Alerts (MPA). 
Specifically, in MPA, when a mis-prediction occurs during pre-training, we use its co-occurrence information to guide several heads of the self-attention modules. These self-attention heads in the Transformer modules are optimized to assign higher attention weights to the words in the input sentence that rarely co-occur with the mis-prediction while assigning lower weights to the other words.
By doing so, the Transformer model is trained to rely less on the dominating co-occurring patterns of mis-predictions while focusing more on the rest of the contextual information.

Empirically we find that such a simple regularization can help pre-training methods quite effectively. We conduct experiments using BERT and ELECTRA~\citep{clark2019electra} as MPA's backbone methods. Our results show that MPA expedites the training processes of BERT and ELECTRA and improves their performances on downstream tasks.
Sensitivity analysis on hyper-parameter configurations shows that MPA improves the backbone methods in a wide range of hyper-parameter settings. The ablation study also illustrates that MPA's effectiveness largely comes from the assigning appropriate attention weights to rare context of mis-predictions.

\section{Using Mis-Predictions as Harm Alerts}
\label{section:main}
\vspace{-2mm}
In the section, we first briefly introduce BERT (in Appendix \ref{appx:preliminaries}) and ELECTRA as preliminaries.
Then we describe the proposed method, Mis-Predictions as Harm Alerts (MPA).
We use ELECTRA as the backbone method to describe MPA for ease of illustration, but it is important to note that MPA can potentially be applied to other Transformer-based language pre-training methods as well. 

\subsection{Preliminaries}
\label{subsection:preliminaries}

\paragraph{ELECTRA} ELECTRA trains two Transformer models in parallel, one smaller BERT as generator (G) and the other normal-sized Transformer as discriminator (D). The training objective of the discriminator is to detect the mis-predicted tokens from the generator. 

The ELECTRA generator's training objective and loss function are the same as those of BERT, as described in the previous section and Equation~\ref{eq:bert-loss}.
Specifically, the input of the generator is a token sequence masked similarly as in BERT pre-training. At every iteration during training, the generator makes predictions at the masked positions of each sequence. Then the generator outputs a new token sequence with tokens at the masked positions replaced by the generator's predictions. The generator's output sequence then would be used as the input of the discriminator. If we denote the generator's output sequence as $\mathbf{x}^r$, then the discriminator's loss function is,

\begin{equation}
    \begin{aligned}
        \mathcal{L}_{D}(\mathbf{x}, \mathbf{x}^{r}) &= \frac{1}{N} (\sum_{t=1}^N-\mathbf{1}(\mathbf{x}_t^{r}=\mathbf{x}_t)\cdot \log D(\mathbf{x}^{r},t) \\
        &-\mathbf{1}(\mathbf{x}_t^{r}\neq \mathbf{x}_t)\cdot\log (1-D(\mathbf{x}^{r},t))).
    \label{eq:electra-loss}
    \end{aligned}
\end{equation}

$N$ is the length of the input sequence $x$. Equation~\ref{eq:electra-loss} is a cross-entropy loss for the binary classification task of ELECTRA's discriminator. With this loss, the discriminator of ELECTRA is trained to detect if the token at each position of the input is a mis-prediction. The final combined loss of ELECTRA with respect to $x$ is,
\vspace{-3mm}
\begin{align}
\mathcal{L}_(\mathbf{x},\mathbf{x}^{m}, \mathbf{x}^{r}) &= \mathcal{L}_G(\mathbf{x}, \mathbf{x}^{m})+\lambda \mathcal{L}_D(\mathbf{x}, \mathbf{x}^{r}),
\label{eq:electra-entire-loss}
\end{align}
$\lambda$ is the hyper-parameter controlling the ratio of the two losses. With this modified task, ELECTRA improves the pre-training efficiency by providing a better sampling efficiency.

\subsection{The Proposed Method}
\label{subsection:main}
\vspace{-2mm}
A mis-prediction usually occurs when there are dominating co-occurring patterns with the mis-prediction in the context that lead to it. 
In this section, we firstly describe how MPA locates such dominating and harmful patterns with the help of mis-predictions.  MPA achieves this goal by preparing beforehand a context matrix $\mathbf{S}$ with word oc-occurrence information. In context matrix $\mathbf{S}$, for a token $i$ and token $j$, if $j$ co-occurs frequently with $i$, $\mathbf{S}_{i,j}$ is close to $1$ and vice versa. Details of constructing $\mathbf{S}$ are in Appendix\ref{appx:context}. 
Then we describe how MPA trains the Transformer model to rely less on such patterns and rely more on the rest information with the help of this context matrix.

\paragraph{Pre-training.} After recording the context information of tokens in the context matrix $S$, we start pre-training with MPA. During pre-training, given an partially-masked input sequence $\mathbf{x}^m$, we firstly forward it to the generator and get $\mathbf{x}^r$, same as ELECTRA. $\mathbf{x}^r$ is the output sequence from the generator, with tokens at the masked positions replaced by the generator's predictions. Then we collect the mis-predicted positions in $\mathbf{x}^r$ as,
\begin{align}
\mathcal{M}_{x} =\{1(x_t^{r} \neq x_t)\cdot t\}_{t=1}^{N_x}.
\label{eq:multi-head}
\end{align}

\begin{figure}[t]
  \centering
  \includegraphics[width=4.1in]{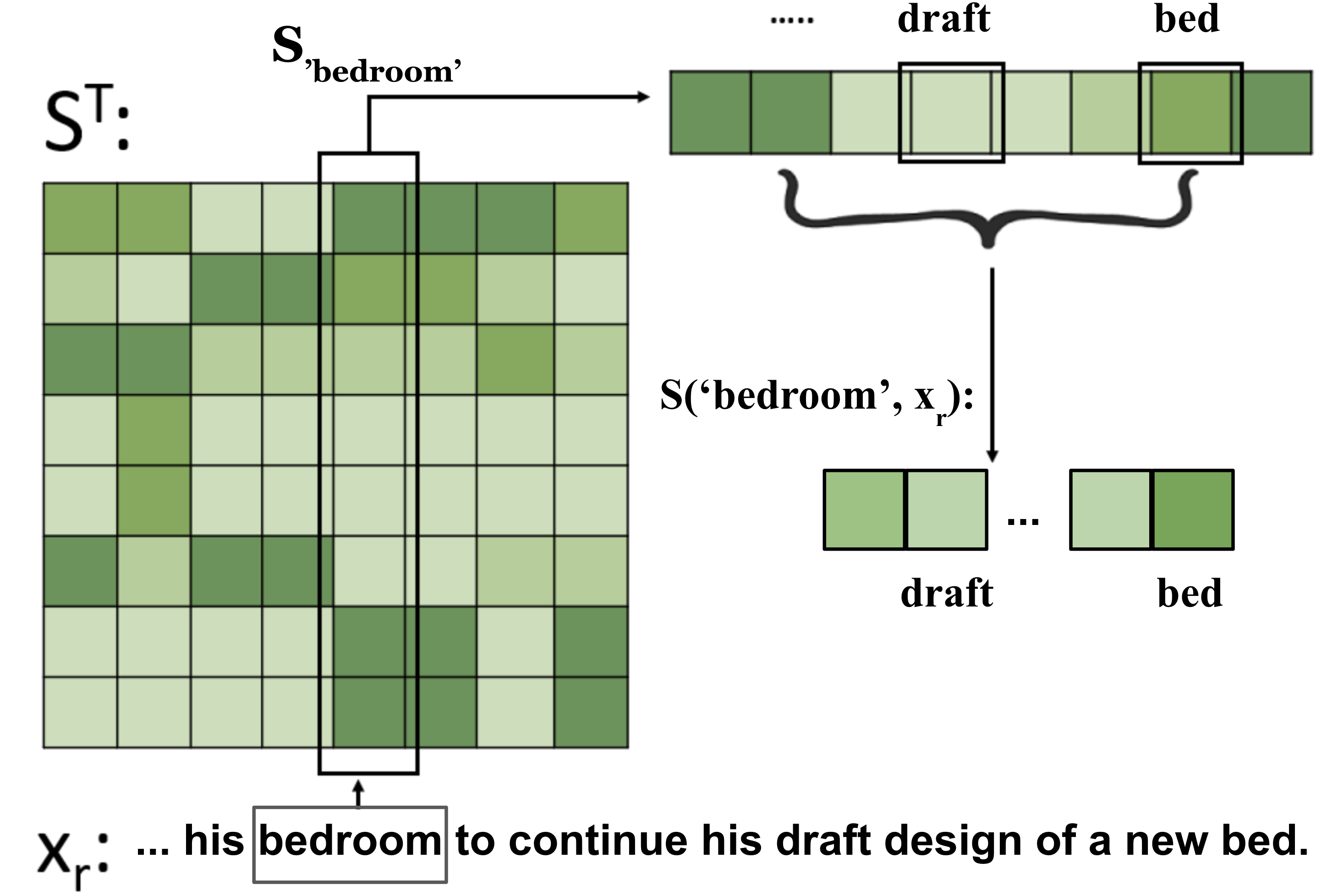}  
  \caption{ Given the input sentence and the mis-prediction ``bedroom'', From the pre-calculated context matrix $\mathbf{S}$, we first fetch $\mathbf{S}_{\text{'bedroom'}}$. $\mathbf{S}_{\text{'bedroom'}}$ consists of the context co-efficients of `is' with all tokens in the vocabulary. Then from $\mathbf{S}_{\text{'bedroom'}}$, MPA fetches the context co-efficients of `bedroom' with all tokens in the sentence $\mathbf{x}_r$, such as `draft' and `bed'. We denote this context co-efficient vector as $S(\text{'bedroom'}, \mathbf{x}_r)$. 
  }
  \label{fig:fetch}
\end{figure}

$N_x$ is the length of the input sequence $\mathbf{x}$. As described in Equation~\ref{eq:attention}, each query-key attention co-efficient $a(i,j)$ is calculated by a scaled dot-product of the query $i$ and the key $j$. We maintain the self-attention modules the same for all the other positions in the input sequence, except the mis-predicted positions in $\mathcal{M}_{x}$. 
For each mis-predicted position $t$ in the input sequence $\mathcal{M}_{x}$, we firstly fetch the pre-calculated context vector of the mis-predicted token $x_t^{r}$ from $\mathbf{S}$. We denote it as $\mathbf{S}_{x_t^r}$.
The vector $\mathbf{S}_{x_t^r}$ consists of the context co-efficients of the mis-prediction $x_t^{r}$ with all tokens in the vocabulary.
Then from $\mathbf{S}_{x_t^r}$, we select context coefficients of mis-prediction $x_t^{r}$ with tokens in the input sentence $\mathbf{x}^r$ and denote the vector as $\mathbf{S}(t, \mathbf{x}^r)$. $\mathbf{S}(t, \mathbf{x}^r)$ is a vector with its dimension equal to the length of $\mathbf{x}^r$. Each element $\mathbf{S}(t, \mathbf{x}^r)_i$ in $\mathbf{S}(t, \mathbf{x}^r)$ is the context coefficient of token pair $(x_t^r, x_i^r)$. Figure~\ref{fig:fetch} describes this process in detail with an example. 

With $\mathbf{S}(t, \mathbf{x}^r)$ prepared for the input sequence $\mathbf{x}^r$ and the mis-predicted position $t$, MPA guides several self-attention heads in the discriminator to focus less on the frequent context and more on the other context of each mis-prediction $x_t^{r}$.
Specifically, at the mis-predicted position $t$, for query $\mathbf{q}_t$, the context-guided self-attention calculates the attention coefficients as, 
\vspace{-2mm}
\begin{align}
\mathbf{g}(\mathbf{q}_t, \mathbf{K}) = \frac{\mathbf{q}_t\mathbf{K}^T}{\sqrt{d_K}}\cdot (1-\mathbf{S}(t, \mathbf{x}^r)).
\label{eq:context-guilded}
\end{align}
\vspace{-1mm}
In this equation, we multiply the original pre-softmax attention co-efficients $\frac{\mathbf{q}_t\mathbf{K}^T}{\sqrt{d_K}}$ at position $t$ with $(1-\mathbf{S}(t, \mathbf{x}^r))$.
Through this way, keys ignored by the attention module could be set a larger weight in $\mathbf{g}(\mathbf{q}_t, \mathbf{K})$ if their context coefficients in $\mathbf{S}(t, \mathbf{x}^r)$ are smaller compared with other tokens in the sentence. Keys at positions of frequent context of the mis-prediction would be set a smaller weight in $\mathbf{g}(\mathbf{q}_t, \mathbf{K})$ since their context coefficients in $\mathbf{S}(t, \mathbf{x}^r)$ is relatively large.

We  then use $\mathbf{g}(\mathbf{q}_t, \mathbf{K})$ as the supervised information to train these self-attention heads at the mis-predicted position $t$. Specifically, we minimize the L-2 loss between $\mathbf{g}(\mathbf{q}_t, \mathbf{K})$ and the original pre-softmax attention weights from the attention module,
\vspace{-3mm}
\begin{align}
\small
\mathcal{L}_A = \frac{1}{N_M}\sum_{t=0}^{N_M}(\frac{\mathbf{q}_t\mathbf{K}^T}{\sqrt{d}}-\mathbf{g}(\mathbf{q}_t, \mathbf{K}))^2
\label{eq:attention_loss_2}
\end{align}
\vspace{-3mm}

$N_M$ is the total number of mis-predictions. Note that we don't back-propergate through $\mathbf{g}(\mathbf{q}_t, \mathbf{K})$. $\mathbf{g}(\mathbf{q}_t, \mathbf{K})$ is only used as supervised information to guide the model to focus more on conflicting context of mis-predictions. Figure~\ref{fig:arch} in Appendix shows with an example the training of self-attention heads with this additional loss at the mis-predicted positions. 
The final loss function of the proposed pre-training method is,
\vspace{-3mm}
\begin{align}
\small
\mathcal{L}_(\mathbf{x},\mathbf{x}^{m}, \mathbf{x}^{r}) &= \mathcal{L}_G(\mathbf{x}, \mathbf{x}^{m})+\lambda \mathcal{L}_D(\mathbf{x}, \mathbf{x}^{r}) + \gamma \mathcal{L}_A (\mathbf{x}, \mathbf{x}^{r}).
\label{eq:total-loss}
\end{align}
\vspace{-1mm}
$\lambda$ and $\gamma$ and hyper-parameters controlling the ratio of the 3 losses. We present the implementation details in Appendix \ref{appx:implementation}.

\section{Experiments}
\label{section:experiments}
\vspace{-2mm}

To verify the efficiency and effectiveness of MPA, we conduct pre-training experiments and evaluate pre-trained models on fine-tuning downstream tasks. 
All codes are implemented with \emph{fairseq} \citep{ott2019fairseq} in \emph{PyTorch} \citep{paszke2017automatic}. All models are run on 8 NVIDIA Tesla V100 GPUs with mixed-precision \citep{micikevicius2017mixed}. We describe our experiment setup in Appendix \ref{appx:exp_setup}

\begin{table}
\small
\caption{Performances of all methods on all downstream tasks. Results of GPT, BERT and ELECTRA are from \citep{clark2019electra} and \citep{levine2020pmi}. The result of SpanBERT is obtained by fine-tuning the released checkpoint from \citep{joshi2019spanbert}. We also reproduce BERT and ELECTRA in our system for fair comparison. We report their results as BERT(ours) and ELECTRA(ours).
MPA outperforms its corresponding backbone methods on every downstream task.
}
\label{tab:general}
\small
\begin{tabular}{lccccc}
\toprule
 & Params & Avg. GLUE  & Avg. SuperGLUE & SQuAD2.0 F1 & SQuAD2.0 EM\\
\hline
GPT-2 & 117 M & 78.8 & -&- &-\\
BERT & 110 M & 82.2 & 66.1 &76.4 &79.6\\
SpanBERT & 110 M & 83.9 &- &77.1 &80.3\\
ELECTRA & 110 M & 85.1 & -& 80.5&83.3\\

\hline
BERT (Ours) & 110 M & 83.0 & 66.3& 76.9& 80.1\\
BERT-MPA & 110 M & $\mathbf{83.7}$ & $\mathbf{67.4}$& $\mathbf{77.5}$& $\mathbf{80.7}$\\
\hline
ELECTRA (Ours) & 110 M & 85.2 & 70.1& 80.2 &83.1\\
\textbf{ELECTRA-MPA} & 110 M & $\mathbf{86.0}$ & $\mathbf{72.2}$ & $\mathbf{83.1}$ & $\mathbf{86.1}$\\
\bottomrule
\end{tabular}
\end{table}

Table~\ref{tab:general} shows the performances of all pre-trained models on every task. We can see that MPA improves the performance of both BERT and ELECTRA on each of the task tested. For fair comparison, we pre-train the two baselines, BERT and ELECTRA, with our system and data and present the results as BERT(ours) and ELECTRA(ours). 
We also present pre-trained models' performances on every individual sub-task of GLUE in Table~\ref{tab:glue} and those of SuperGLUE in Table~\ref{tab:superglue} in Appendix. Results show that MPA can outperform both of the backbone methods in the majority of sub-tasks.  MPA also has considerable performance improvements on tasks requiring more complex semantic understanding and reasoning, such as MNLI, RTE, WiC and WSC.
We also present the average GLUE score curves during the pre-training in Figure~\ref{fig:curve} in Appendix. It shows that MPA outperforms its backbone methods throughout the entire pre-training. It also shows that MPA can reach the same average GLUE scores with fewer numbers of iterations compared with both baselines. The number of iterations required for MPA-ELECTRA is $40\%$ less than that of ELECTRA when reaching ELECTRA's final performance. Since MPA-ELECTRA's running time per iteration is identical with ELECTRA, MPA-ELECTRA's training time is also $40\%$ less than that of ELECTRA. 

Finally we present our hyper-parameter analysis and ablation study in Appendix \ref{appx:ablation_study}

\section{Conclusion and Future Work}
\label{section:conclusion}
\vspace{-2mm}
In this work, we propose MPA (Mis-Predictions as Harm Alerts) to improve language pre-training methods. In MPA, we train several heads of Transformer's attention modules to rely less on the dominating co-occurring patterns with mis-predictions and to focus more on the rare context of them. Experimental results show that MPA can improve both BERT and ELECTRA on the majority of GLUE downstream tasks. MPA can also make the pre-training of BERT and ELECTRA more efficient. 
The ablation study also confirms that MPA's effectiveness largely comes from the rare context of mis-predictions. 



\bibliography{example_paper}
\bibliographystyle{plain}

\clearpage
\appendix
\section{Appendix}
\subsection{Preliminaries}
\label{appx:preliminaries}
\paragraph{BERT} We first use BERT as an example to introduce the basic model architecture and training objectives of language pre-training methods. BERT (\textbf{B}idirectional \textbf{E}ncoder \textbf{R}epresentation from \textbf{T}ransformers) is a multi-layer bidirectional Transformer encoder model, which takes the combination of semantic (token embeddings) and ordering information (positional embeddings) of a sequence of words as input, and outputs a sequence of contextualized token representations of the same length. 

Each Transformer layer consists of a self-attention sub-layer and a position-wise feed-forward sub-layer, with a residual connection~\citep{he2016deep} and layer normalization~\citep{ba2016layer} applied after every sub-layer. The self-attention sub-layer is referred to as "Scaled Dot-Product Attention" in~\citep{vaswani2017attention}, which produces its output by calculating the scaled dot products of \emph{queries} and \emph{keys} as the coefficients of the \emph{values}, i.e.,
\begin{align}\label{eq:attention}
\mathbf{a}(\mathbf{q},\mathbf{K}) &= \frac{\mathbf{q}\mathbf{K}^T}{\sqrt{d_K}}\\
\text{Attention}(\mathbf{q},\mathbf{K},\mathbf{V})&=\text{Softmax}(\mathbf{a}(\mathbf{q}, \mathbf{K}))\mathbf{V}.
\end{align}
$\mathbf{q}$ (Query),
$\mathbf{K}$ (Key), $\mathbf{V}$ (Value) are the hidden states outputted from the previous layer and $d_K$ is the dimension of the hidden representations. Among them, $\mathbf{q}$ is the hidden state vector for the query token.
$\mathbf{K}$ and $\mathbf{V}$ are matrices, each row of which is the hidden state vector for a key/value position. $\mathbf{a}(\mathbf{q},\mathbf{K})$ is $\mathbf{K}$'s attention coefficient vector with the dimension the same as the number of keys in $\mathbf{K}$. Each Transformer layer further extends the self-attention layer described above to a multi-head version in order to jointly attend to information from different representation subspaces. The multi-head self-attention sub-layer works as follows,
\begin{align}
&\text{Multi-head}(\mathbf{q},\mathbf{K},\mathbf{V}) = \text{Concat} (\text{head}_1,\cdots,\text{head}_H)\mathbf{W}^O \\
&\text{head}_k = \text{Attention}(\mathbf{q}\mathbf{K}_k^Q, \mathbf{K}\mathbf{W}_k^K,\mathbf{V}\mathbf{W}_k^V),
\label{eq:multi-head-bullshit}
\end{align}
where $\mathbf{W}_k^Q\in \mathbb{R}^{d \times d_K}, \mathbf{W}_k^K\in \mathbb{R}^{d \times d_K}, \mathbf{W}_k^V\in \mathbb{R}^{d\times d_V}$ are projection matrices. $H$ is the number of heads. $d_K$ and $d_V$ are the dimensions of the key and value separately.

Following the self-attention sub-layer, there is a position-wise feed-forward (FFN) sub-layer, which is a fully connected network applied to every position identically and separately. The FFN sub-layer is usually a two-layer feed-forward network with a ReLU activation function in between. Given vectors $\{\mathbf{h}_1, \ldots, \mathbf{h}_n\}$, a position-wise FFN sub-layer transforms each $\mathbf{h}_i$ as $\text{FFN}(\mathbf{h}_i) = \sigma(\mathbf{h}_i\mathbf{W}_1+\mathbf{b}_1) \mathbf{W}_2 + \mathbf{b}_2$, where $\mathbf{W}_1, \mathbf{W}_2, \mathbf{b}_1$ and $\mathbf{b}_2$ are parameters.

BERT uses the Transformer model as its backbone architecture and trains the model parameters with the Masked Language Model (MLM) task on large text corpora.
In the MLM task, $15\%$ of the positions in an input sentence, $\mathbf{x}$, are randomly selected, and each of the selected positions might be replaced by the special token \texttt{[MASK]}, replaced by randomly chosen tokens, or remain the same. The objective of BERT pre-training is to predict tokens at the selected positions correctly given the masked sentence. If we denote the masked version of $\mathbf{x}$ as $\mathbf{x}^{m}$, then BERT's loss function is
\begin{align}
\mathcal{L}(\mathbf{x}, \mathbf{x}^{m}) &= \frac{1}{N_x} (\sum_{t=1}^{N_{x}} -\log p_{\theta}(x_t|\mathbf{x}^m)),
\label{eq:bert-loss}
\end{align}

where $N_x$ is the total number of masked positions in the input token sequence $x$ and $\theta$ is the trainable parameters of the Transformer. $x_t$ is the ground-truth token at a masked position $t$ in $\mathbf{x}$. Specifically, for every masked position $t$, BERT takes its output contextual representation at the same position $t$ to make a prediction.

As this task requires no human labeling effort, large scale data corpus is usually used to train the model. Empirically, the trained model, served as a good initialization, significantly improves the performance of downstream tasks.

\subsection{Calculating the Context Matrix in MPA}
\label{appx:context}
In this context matrix, we record word/token-level co-occurrence information from the pre-training corpora before the pre-training. 
To record the token-level co-occurrence information, we firstly calculate the total numbers of co-occurrence for all token pairs in the vocabulary. If we denote the vocabulary of the training corpus as $V=\{w_1, w_2, \cdots, w_V\}$, then we calculate a matrix $C$, the element $C_{i,j}$ in which records the total number of times that token $w_i$ and $w_j$ occur together. 
The co-occurrence matrix $C$ needs to be normalized so that it is irrelevant with the tokens' own occurrences. 
Without this process, the co-occurrence matrix would be dominant by token occurrences. Rare tokens would have low co-occurrence co-efficients with all other tokens. Common tokens would have high co-occurrence co-efficients with all other tokens. 
In order to prevent it from happening, we normalize $C$ as, 

\begin{align}
\mathbf{C}^{normed}_{i,j} =\frac{\mathbf{C}_{i,j}}{(\sum_{z=1}^{V}\mathbf{C}_{i,z})\cdot(\sum_{z=1}^V\mathbf{C}_{j,z})},
\label{eq:c-normed}
\end{align}

$\sum_{z=1}^{V}\mathbf{C}_{i,z}$ and $\sum_{z=1}^{V}\mathbf{C}_{j,z}$ are the total number of occurrences of token $i$ and $j$ respectively. Then we calculate the final context matrix $S$. The $i$-th row of $\mathbf{S}$ is the final context co-efficient vector of token $i$,
\begin{align}
\mathbf{S}_{i} = \frac{\mathbf{C}^{normed}_{i}-\min(\mathbf{C}^{normed}_i)}{\max(\mathbf{C}^{normed}_i)- \min(\mathbf{C}^{normed}_i)}.
\label{eq:s-i}
\end{align}

In this equation, every row $\mathbf{C}^{normed}_{i}$ is normalized to the range of $[0,1]$. Without this process, the pre-training of MPA will be unstable and the model can diverge very easily.
For a token $i$ and token $j$, if $j$ co-occurs frequently with $i$, $\mathbf{S}_{i,j}$ is close to $1$ and vice versa. 
With this context matrix $\mathbf{S}$,
for a mis-prediction $i$ at pre-training and a token $j$ in this input sentence, if $\mathbf{S}_{i,j}$ is large, MPA will train several self-attention heads in the model to output a smaller attention co-efficient $a(i,j)$.
Through this way, MPA can rely less on the context that co-occurs frequently with the mis-prediction, and focus more on the rest context. 

\begin{figure}[t]
  \centering
  \includegraphics[width=4.1in]{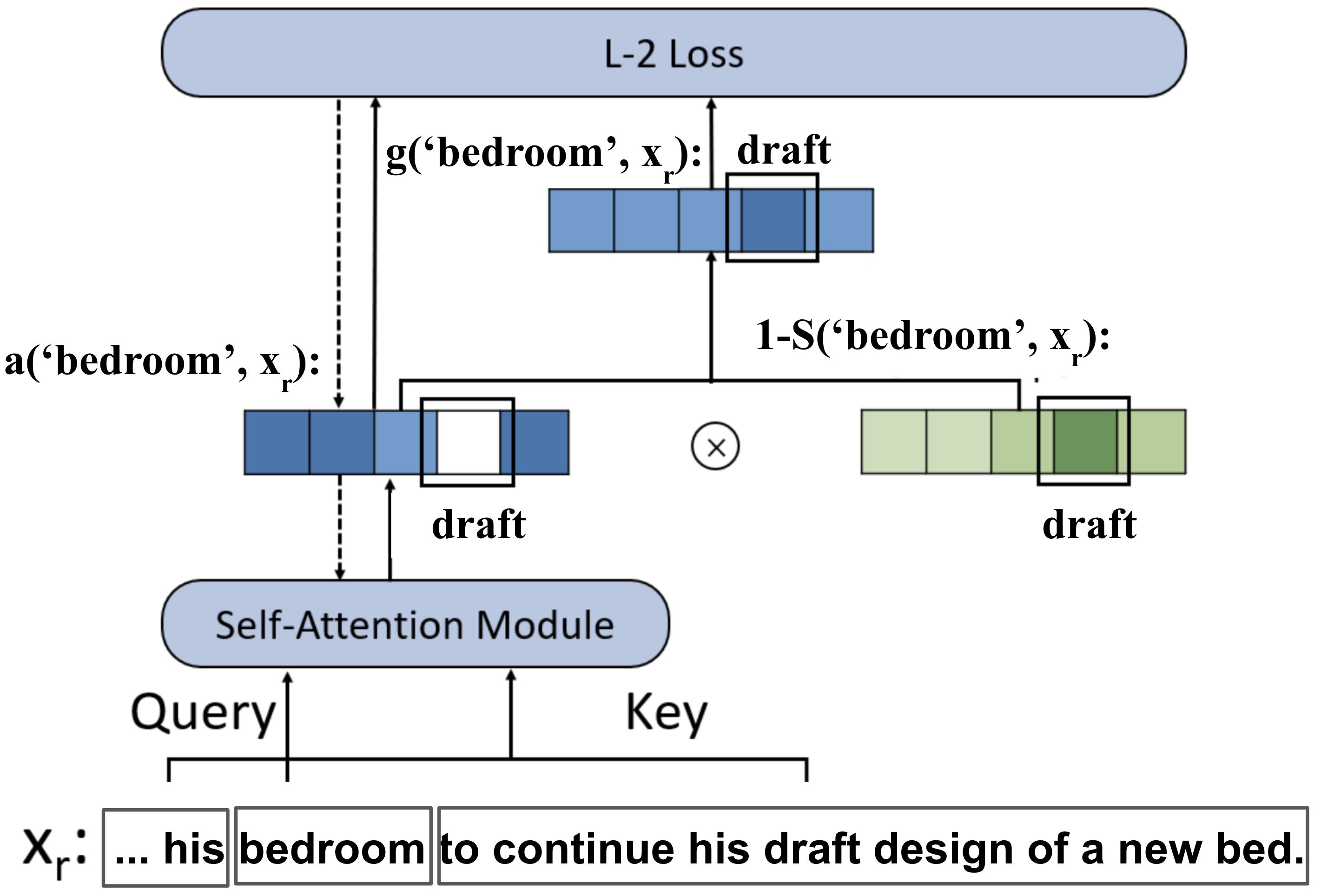}  
  \caption{The training of one self-attention head at the mis-predicted positions in MPA. The blue vector in the left side is the original pre-softmax attention co-efficient vector $a(\text{'bedroom'},\mathbf{x}_r)$. The original attention co-efficient at the 'draft' position in $a(\text{'bedroom'},\mathbf{x}_r)$ is smaller than the rest. By multiplying it with $1-S(\text{'bedroom'},\mathbf{x}_r)$ (the green vector in the right side), we get $g(\text{'bedroom'},\mathbf{x}_r)$. In  $g(\text{'bedroom'},\mathbf{x}_r)$, the attention coefficient at the 'draft' position is larger than the rest. 
  The dashed arrows represent the backward pass and the firm arrows represent the forward pass.}
  \label{fig:arch}
  \vspace{-3mm}
\end{figure}

\subsection{Implementation Details}
\label{appx:implementation}
In the previous sub-section, we described how MPA trains one-layer one-head attention module at mis-predicted positions. While Transformer has multiple layers of multi-head attention modules, we denote the number of self-attention layers trained with MPA as $l$ (in a bottom-up order), and the number of attention heads at every layer trained with MPA as $h$. We set $h$ as $3$ and $l$ as $5$ for ELECTRA-MPA. We set $h$ as $3$ and $l$ as $5$ for BERT-MPA. We also analyze the behaviour of MPA when $h$ and $l$ vary under Section~\ref{section:experiments}. When we choose the value of $h$ and $l$, we not only decide according to average GLUE score, but also take into consideration the model's performance on more complex sub-tasks such as MNLI and QNLI. For the pre-calculated context matrix $\mathbf{S}$, the size of $\mathbf{S}$ would be too big to fit into the GPU memory together with the model if we include co-occurrence information for every word pair. Therefore after experimenting we set $\mathbf{S}$'s vocabulary size to be $5000$. We only record the co-occurrence information for words with top $5000$ frequencies in the training set. For words not in the vocabulary of $\mathbf{S}$, we train the attention modules the same as MPA's backbone method even if they are mis-predictions. Increasing the vocabulary of $\mathbf{S}$ to be larger than $5000$ brings minor performance improvements compared with the memory overhead.

\subsection{Experimental Setup}
\label{appx:exp_setup}

To show the wide adaptability of MPA, we use both ELECTRA \citep{clark2019electra} and BERT \citep{devlin2018bert} as the backbone language pre-training methods and implement MPA on top of them. In the previous section, we have described MPA on top of ELECTRA. Similarly, the implementation of MPA on top of BERT requires a small generator.
In MPA-BERT, we also train a small generator in parallel with BERT and abandon the generator at fine-tuning, same as ELECTRA. The size of the generator is also the same as that of the generator in ELECTRA. In MPA-BERT, when training BERT with its original Masked Language Modeling task, we also train several of its attention heads to rely less on the frequent context of mis-predictions from the generator as described in Equation~\ref{eq:attention_loss_2}. Note that in BERT-MPA, we don't replace the the masked positions of the BERT inputs with the generator's predictions. The predictions from the generator are only used for minimizing Equation~\ref{eq:attention_loss_2} (the MPA loss) and BERT's original MLM task remains unchanged in BERT-MPA. 

We fine-tune the pre-trained models on GLUE (\textbf{G}eneral \textbf{L}anguage \textbf{U}nderstanding \textbf{E}valuation) \citep{DBLP:journals/corr/abs-1804-07461}, SuperGLUE \citep{wang2019superglue} and SQuAD 2.0~\citep{rajpurkar2018know} to evaluate the performance of the pre-trained models. The details of downstream fine-tuning are described below.

\paragraph{Fine-tuning on Downstream Tasks.} 
For GLUE fine-tuning, We follow previous work to use eight tasks of GLUE, including CoLA, RTE, MRPC, STS, SST, QNLI, QQP, and MNLI. 
For evaluation metrics on the sub-tasks, we report Matthews correlation for CoLA, Pearson correlation for STS-B, and accuracy for other tasks.
SuperGLUE also consists of 8 natural language understanding tasks, including question answering (MultiRC, BoolQ, ReCoRD) \citep{zhang2018record,clark2019boolq,khashabi2018looking}, textual
entailment (SuperRTE, CB) \citep{dagan2005pascal,clark2019boolq}, coreference resolution (WSC) \citep{levesque2012winograd}, word sense disambiguation (WiC) \citep{pilehvar2019wic}, and
causal reasoning (COPA) \citep{roemmele2011choice}. We adopt the
same evaluation metrics as ~\citep{wang2019superglue}. For SuperRTE, BoolQ, COPA, WiC and WSC, we report the classification accuracy. For CB, we report F1/Accuracy of the classification. For MultiRC, we report F1 over all answer options ($F1_a$) and exact match (EM) of each question's set of answers. For ReCoRD, we report F1/EM.
SQuAD~\citep{rajpurkar2016squad, rajpurkar2018know} has served as a major question answering benchmark for pretrained models. It provides a paragraph of context and a question, and the task is to answer the
question by extracting the relevant span from the context. We focus on the latest more challenging variant, SQuAD2.0~\citep{rajpurkar2018know}, some questions in which are not answered in the
provided context.
We use the same optimizer (Adam) with the same hyper-parameters as in pre-training. Following previous work, we search the learning rates during the fine-tuning for each downstream task. The details are listed in Table~\ref{tab:ds_space}. For fair comparison, we do not apply any published tricks for fine-tuning. Each configuration is run five times with different random seeds, and the \emph{average} of these five results on the validation set is calculated as the final performance of one configuration.

\begin{table*}[]
\small
\centering
\caption{Performance of different models on SuperGLUE sub-tasks.For every pre-trained model, we fine-tune each sub-task with 5 random seeds and report the average. Results show that MPA outperforms backbone methods on the majority of individual tasks. }
\label{tab:superglue}
\begin{tabular}{lcccccccc|c}
\toprule
        & SuperRTE& BoolQ & CB & COPA & MultiRC & ReCoRD & WiC & WSC & Avg.  \\
\hline
\hline
BERT (Ours)&  67.2 & 72.9 & 83.2/80.3& 67.0 & 68.5/20.3&  \textbf{65.5/64.1} & 68.2 & 64.3 & 66.3\\
\hline
BERT-MPA  &\textbf{69.8} & \textbf{74.2} & \textbf{83.7/80.4} & \textbf{67.0} & \textbf{69.1/22.4} & 64.1/63.9 & \textbf{69.6} &\textbf{66.7}   & \textbf{67.4}\\
\hline
\bottomrule
ELECTRA(Ours)  &72.3 & 78.3 & 86.6/87.3 & \textbf{69.6} & \textbf{68.7/20.7}& 68.8/67.3& 70.3& 70.5& 70.1\\
\hline
ELECTRA-MPA &\textbf{75.3} & \textbf{81.2} & \textbf{91.4/90.7} &69.0& 68.5/21.4& \textbf{69.8/68.5} & \textbf{73.8}& \textbf{72.5}  & \textbf{72.2}\\
\hline

\bottomrule
\end{tabular}
\end{table*}

\begin{table*}[]
\small
\centering
\vspace{-2mm}
\caption{Performance of different models on GLUE sub-tasks. For every pre-trained model, we fine-tune each sub-task with 5 random seeds and report the average. Results show that MPA outperforms backbone methods on the majority of individual tasks. }

\label{tab:glue}
\begin{tabular}{lcccccccc|c}
\toprule
        & MNLI& QNLI & QQP & SST & CoLA & MRPC & RTE & STS & Avg.  \\
\hline
\hline
BERT (Ours)  & 84.93 & \textbf{91.34} & 91.04 & 92.88 & 55.19 & 88.29 & 68.61 & \textbf{89.43} & 82.96 \\
\hline
BERT-MPA  & \textbf{84.94} & 90.88 & \textbf{91.06} & \textbf{93.37} &\textbf{ 61.21} & \textbf{89.73} & \textbf{70.57} & 87.81 & \textbf{83.70} \\
\hline
\bottomrule
ELECTRA(Ours)  &86.93 &92.50 &91.57 &93.01 &\textbf{67.58} &90.29 &70.12 & 89.98 &85.22 \\
\hline
ELECTRA-MPA &\textbf{87.08} & \textbf{92.76} & \textbf{91.70} &\textbf{93.79} & 67.42 & \textbf{91.78} & \textbf{73.17} & \textbf{90.16} & \textbf{86.02} \\
\hline

\bottomrule
\end{tabular}
\vspace{-2mm}
\end{table*}

\paragraph{Data Corpus and Pre-training Tasks.} 
Following BERT \citep{devlin2018bert}, we use English Wikipedia corpus and BookCorpus \citep{moviebook} for pre-training. By combining these two datasets, we obtain a corpus with roughly 16GB in size, similar to \citep{devlin2018bert}. We also conduct a series of pre-processing of the language data following existing works: segmenting documents into sentences by Spacy\footnote{\url{https://spacy.io}}, normalizing, lower-casing, tokenizing the texts by Moses decoder \citep{Koehn2007MosesOS}, and finally, applying byte pair encoding (BPE) \citep{DBLP:journals/corr/SennrichHB15} with the vocabulary size set as 32,768. 
We use \textit{masked language modeling} as the objective of BERT pre-training and \textit{replaced token detection} for ELECTRA pre-training. We remove the next sentence prediction task and use \textit{FULL-SENTENCES} mode to pack sentences as suggested in RoBERTa \citep{liu2019roberta}. 
In Masked Language Modeling of BERT, the masked probability is set to $0.15$.  After masking, we replace 80\% of the masked positions with \texttt{[MASK]}, 10\% by randomly sampled words, and keep the remaining 10\% unchanged. 
In Replaced Token Detection of ELECTRA, the masked probability of the generator is set to $0.15$ and all masked positions are replaced with \texttt{[MASK]}. We report all the details of model architectures and hyper-parameters of both backbone methods and MPA in the following paragraph.

\begin{table*}[]
\small
\centering
\caption{Experimental results on the sensitivity of MPA's hyper-parameter $l$, $h$ and $\gamma$.}
\label{tab:hyper-result}
\addtolength{\tabcolsep}{-1pt}    
\begin{tabular}{l|ccc|cc|cc|cc|cc|c}
\toprule
& \multicolumn{11}{c}{ELECTRA-MPA} \\ \hline
    Run \# & R1 & R2 & R3 & R4& R5 & R6 & R7 & R8 & R9 & R10 & R11 & R12\\ \hline
    
     $l$  & 1     & 1     & 1      & 3     & 3     & 5     & 5     & 7     & 7     & 1    & 1   & 0\\
     $h$  & 3     & 5     & 7      & 3     & 5     & 3     & 5     & 3     & 5     & 5    & 5   & 0\\
$\gamma$  & 1.0   & 1.0   & 1.0    & 1.0   & 1.0   & 1.0   & 1.0   & 1.0   & 1.0   &0.1   & 10  & 0\\ \hline
Avg. GLUE & 85.57 & 85.73 & 84.83  & 85.67 & 85.46 & 86.02 & \textbf{86.08} & 85.52 & 84.48 &85.32 & 84.88 &  85.22    \\\bottomrule
    \toprule
     & \multicolumn{11}{c}{BERT-MPA} \\ \hline
     Run \# & R13 & R14 & R15 & R16 & R17 & R18 &R19 & R20 & R21 & R22 & R23 & R24\\ \hline
     
     $l$  & 1     & 1     & 1     & 3     & 3     & 5    & 5     & 7     & 7    & 1    & 1    & 0\\
     $h$  & 3     & 5     & 7     & 3     & 5     & 3    & 5     & 3     & 5    & 5    & 5    & 0\\
$\gamma$  & 1.0   & 1.0   & 1.0   & 1.0   & 1.0   & 1.0  & 1.0   & 1.0   & 1.0  &0.1   & 10   & 0\\ \hline
Avg. GLUE & 83.44 & 83.39 & 82.90 & 83.33 & 83.30 &\textbf{83.69} & 83.61 & 83.42 &82.24 &83.19 & 82.83 & 82.96\\
\bottomrule
\end{tabular}
\addtolength{\tabcolsep}{2.2pt}
\end{table*}

\paragraph{Model architecture and  hyper-parameters. }
We conduct experiments on ELECTRA-Base\citep{clark2019electra} and BERT-Base (110M parameters)~\citep{devlin2018bert}. BERT consists of 12 Transformer layers for the base model. For each layer, the hidden size is set to 768 and the number of attention head ($H$) is set to 12. The architecture of the discriminator of ELECTRA is the same as BERT-Base. The size of the generator is $1/3$ of the discriminator. We use the same pre-training hyper-parameters for all experiments. All models are pre-trained for 1000$k$ steps with batch size 256 and maximum sequence length 512.
We use Adam \citep{DBLP:journals/corr/KingmaB14} as the optimizer, and set its hyperparameter $\epsilon$ to 1e-6 and $(\beta1, \beta2)$ to (0.9, 0.98). The peak learning rate is set to 1e-4 with a 10$k$-step warm-up stage. After the warm-up stage, the learning rate decays linearly to zero. We set the dropout probability to 0.1 and weight decay to 0.01. 
There are three additional hyper-parameters for MPA, the number of MPA attention layers $l$, the number of MPA attention heads $h$ in each MPA Transformer layer, and the weight $\gamma$ of loss $L_A$. The main results of BERT-MPA we report come from setting $l$ as 5, $h$ as 3, $\gamma$ as 1. The main results of ELECTRA-MPA we report come from setting $l$ as 5, $h$ as 3, $\gamma$ as 1. All hyper-parameter configurations are reported in Table~\ref{tab:ds_space}. We also have a separate sub-section to present and analyze MPA's behaviour at different hyper-parameter settings.
\begin{table}[t]
\centering
\caption{Hyper-parameters for the pre-training and fine-tuning on all language pre-training methods, include both backbone methods and MPAs.} \label{tab:ds_space}
\begin{tabular}{lcc}
\toprule
& Pre-training & Fine-tuning \\ \hline
\textbf{Max Steps} & 1$M$ & - \\
\textbf{Max Epochs} & - & 5 or 10\\ 
\textbf{Learning Rate} & 1e-4 & \{(1,2,3,4,5)$\times$1e-5\}  \\ 
\textbf{Batch Size} & 256 & 32 \\ 
\textbf{Warm-up Ratio} & 0.01 & 0.06 \\ 
\textbf{Sequence Length} & 512 & 512 \\
\textbf{Learning Rate Decay} & Linear & Linear \\ 
\textbf{Adam $\epsilon$} &  1e-6 & 1e-6 \\ 
\textbf{Adam ($\beta_1$, $\beta_2$)} &  (0.9, 0.98) & (0.9, 0.98) \\ 
\textbf{Dropout} & 0.1 & 0.1 \\ 
\textbf{$\lambda$ of ELECTRA} & 50 & - \\
\textbf{Weight Decay} & 0.01 & 0.01 \\ \hline
\textbf{$l$ of BERT-MPA} & 5 & - \\ 
\textbf{$h$ of BERT-MPA} & 3 & - \\ 
\textbf{$\gamma$ of BERT-MPA} & 1 & - \\ 
\textbf{$l$ of ELECTRA-MPA} & 5 & - \\ 
\textbf{$h$ of ELECTRA-MPA} & 5 & - \\ 
\textbf{$\gamma$ of ELECTRA-MPA} & 1 & - \\ 
\bottomrule
\end{tabular}
\end{table}

\vspace{-3mm}
\subsection{Hyper-parameter Analysis and Ablation Study}
\vspace{-3mm}
\label{appx:ablation_study}
We investigate how MPA performs when the hyper-parameters, $\gamma$, $h$ and $l$ vary, and present the results in Table~\ref{tab:hyper-result}.
First, MPA outperforms its backbone methods on the majority of hyper-parameter settings. When $\gamma$ (the coefficient of the MPA loss) varies (R2, R10, R11 for BERT and R14, R22, R23 for ELECTRA), MPA achieves the best performance  when $\gamma$ is set to $1.0$ on both BERT and ELECTRA. When $\gamma$ is $0.1$, MPA is slightly better than the backbone methods on BERT and ELECTRA. When $\gamma$ is set to $10$, MPA performs worse than backbone methods(R12 and R24) on both cases. The reason is that forcing the Transformer model to completely not focus on frequent co-occurrences with mis-predictions can be harmful. 
Rare context in input sentences of mis-predictions should indeed be used to fix the mis-predictions. However, it doesn't mean that the other context in the input sentence is less important. Recall the example sentence ``He went back to his \underline{bedroom} to continue his draft design of a new bed'', if the model only focuses on ``darft design'' and ignores the rest of the words such as ``his'',  it could lead to other mis-predictions. Therefore, the MPA loss can not have an overly large $\gamma$ compared with the original loss of the model. By the same logic, attention modules trained with MPA should not be too ``wide'' or too ``tall'' either. This is verified by the experimental results. We can see that when $h$ reaches 7 or $l$ reaches 7, MPA starts to perform worse than the back-bone methods. 
In order to further verify that it is indeed the larger attention weights of rare context of mis-predictions that lead to better performance, we conduct ablation study by modifying MPA.
\paragraph{Ablation Study}
In order to verify that MPA's effectiveness indeed comes from 1), using mis-predictions as harm alerts instead of using other tokens; 2) training the model to rely less on frequently co-occurring patterns with mis-predictions, we conduct two ablation studies. To confirm the first point, we apply MPA on the ground-truth words at the mis-predicted positions. In other words, we train the model to rely less on frequently co-occurring patterns with the ground-truth word at the mis-predicted position (focusing more on rare context of 'study' instead of 'bedroom' for sentence 'he went back to his \underline{bedroom} to continue his draft design of a new bed'). We denote this method as MPA-Ground. Results are shown in Table~\ref{tab:ablation}.

\begin{wrapfigure}{r}{0.41\textwidth}
  \includegraphics[width=0.44\textwidth]{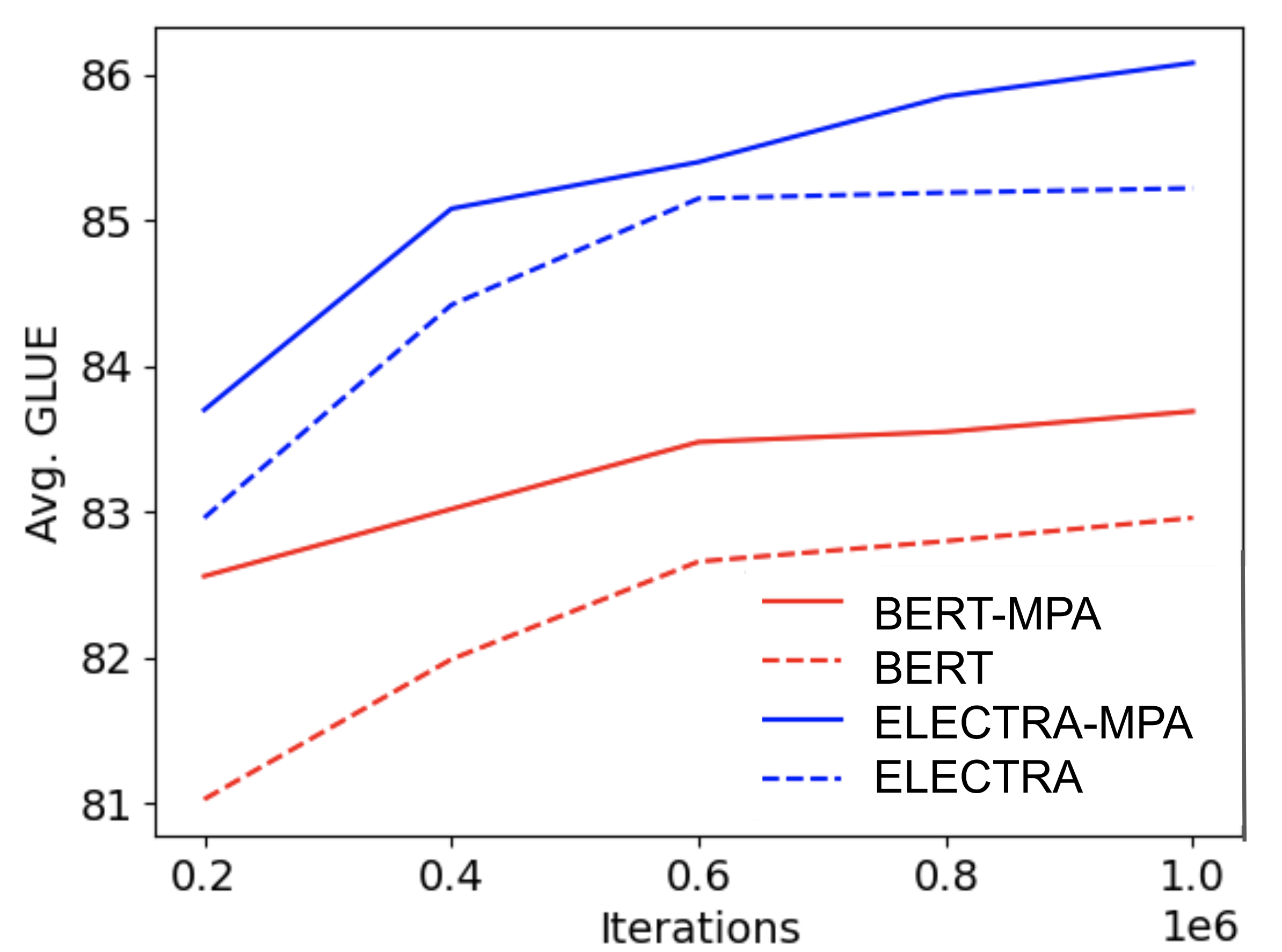}  
  \caption{The curves of average GLUE score for all models trained under the BERT setting and ELECTRA setting. The figure shows that MPA expedites both of the backbone methods.}
  \label{fig:curve}
\end{wrapfigure}

We can see that MPA-Ground outperforms the baseline slightly but performs significantly worse than MPA. It firstly indicates that training Transformer to focus more on rare word-word co-occurrences to a proper extent, can be beneficial for language pre-training methods. The potential reason is that due to the long tail problem of language corpus~\citep{larson2010introduction}, rare word combinations are very under-trained. However, its performances are still far behind MPA. It indicates that although it is possible that MPA's effectiveness could come from its focus on rare word combinations to a minor extent, the main benefits of MPA still comes from mis-predictions as harm alerts.

Second, we would like to verify that the co-occurrence information in $\mathbf{S}$ that varies for every word is important to MPA as well. To confirm this,
we set $S(q_t, K)$ to be constant (optimal constant value $0.8$ for BERT and $0.9$ for ELECTRA ) for every mis-prediction. 
Since we calculated $S$ for the most frequent $5000$ words in the vocabulary, this method will actually train the model to focus equally less on the most frequent $5000$ words. 
We denote this method as MPA-Constant. From Table~\ref{tab:ablation}, we can see that MPA-Constant with an optimal constant value, slightly outperforms the baseline but performs significantly worse than MPA. It indicates that training the model to focus more on rare words to a constant degree helps slightly, but the individual co-occurrence coefficients lead to the main performance gain.

\begin{table}
\caption{Average GLUE scores of ablation methods.}
\centering
\label{tab:ablation}
\begin{tabular}{lccc}
\toprule
 & Params & Avg. GLUE  \\
\hline

\hline
BERT (Ours) & 110 M & 83.0 \\
BERT-MPA & 110 M & $\mathbf{83.7}$  \\
BERT-MPA-Constant & 110 M & 83.1  \\
BERT-MPA-Ground & 110 M & 83.1  \\
\hline
ELECTRA (Ours) & 110 M & 85.2 \\
ELECTRA-MPA & 110 M & $\mathbf{86.1}$ \\
ELECTRA-MPA-Constant & 110 M & 85.4 \\
ELECTRA-MPA-Ground & 110 M & 85.5 \\
\bottomrule
\end{tabular}
\end{table}

\subsection{GLUE and SuperGLUE results with error bars.}
In the section we present the margin of error with Confidence Level $95\%$ of the results of MPA and its backbone methods at each sub-task in GLUE and SuperGLUE. GLUE results are shown in Table \ref{tab:glue-std}  and SuperGLUE in Table \ref{tab:superglue-std}.

\begin{table*}[]
\small
\vspace{-2mm}
\caption{Performance of different models on GLUE sub-tasks. For every pre-trained model, we fine-tune each sub-task with 5 random seeds and report the average. We also report the margin of error with Confidence Level $95\%$. }

\label{tab:glue-std}
\begin{tabular}{lcccccccc}
\toprule
        & MNLI& QNLI & QQP & SST & CoLA & MRPC & RTE & STS \\
\hline
\hline
BERT (Ours)  & 84.9$\pm$0.09 & \textbf{91.3}$\pm$0.17 & 91.0$\pm$0.07 & 92.9$\pm$0.15 & 55.2$\pm$0.63 & 88.3$\pm$0.94 & 68.6$\pm$0.74 & \textbf{89.4}$\pm$0.42 \\
\hline
BERT-MPA  & \textbf{84.9}$\pm$0.11 & 90.9$\pm$0.27 & \textbf{91.1}$\pm$0.16 & \textbf{93.4}$\pm$0.09 &\textbf{ 61.2}$\pm$0.51 & \textbf{89.7}$\pm$0.88 & \textbf{70.6}$\pm$0.81 & 87.8$\pm$0.59  \\
\hline
\bottomrule
ELECTRA(Ours)  &86.9$\pm$0.07 &92.5$\pm$0.11 &91.6$\pm$0.17 &93.0$\pm$0.08 &\textbf{67.6}$\pm$0.28 &90.3$\pm$0.59 &70.1$\pm$0.98 & 90.0$\pm$0.36  \\
\hline
ELECTRA-MPA &\textbf{87.1}$\pm$0.13 & \textbf{92.8}$\pm$0.35 & \textbf{91.7}$\pm$0.19 &\textbf{93.8}$\pm$0.07 & 67.4$\pm$0.24 & \textbf{91.8}$\pm$0.41 & \textbf{73.2}$\pm$0.56 & \textbf{90.2}$\pm$0.29  \\
\hline

\bottomrule
\end{tabular}
\vspace{-2mm}
\end{table*}

\begin{table*}[]
\small
\centering
\caption{Performance of different models on SuperGLUE sub-tasks.For every pre-trained model, we fine-tune each sub-task with 5 random seeds and report the average. We also report the margin of error with Confidence Level $95\%$. }
\label{tab:superglue-std}
\begin{tabular}{lcccccccc}
\toprule
        & SuperRTE& BoolQ & CB & COPA & MultiRC & ReCoRD & WiC & WSC \\
\hline
\hline
BERT (Ours)&  67.2$\pm$1.28 & 72.9$\pm$0.57 & 83.2$\pm$0.09 & 67.0$\pm$0.00 & 68.5$\pm$0.18 &  \textbf{65.5}$\pm$0.08 & 68.2$\pm$0.17 & 64.3$\pm$0.19 \\
\hline
BERT-MPA  &\textbf{69.8}$\pm$1.71 & \textbf{74.2}$\pm$0.91 & \textbf{83.7}$\pm$0.08 & \textbf{67.0}$\pm$0.00 & \textbf{69.1}$\pm$0.37 & 64.1$\pm$0.16 & \textbf{69.6}$\pm$0.56 &\textbf{66.7}$\pm$0.14 \\
\hline
\bottomrule
ELECTRA(Ours)  &72.3$\pm$1.04 & 78.3$\pm$0.38 & 86.6$\pm$0.15 & \textbf{69.6}$\pm$0.49 & \textbf{68.7}$\pm$0.26 & 68.8$\pm$0.13 & 70.3$\pm$0.38 & 70.5$\pm$0.15\\
\hline
ELECTRA-MPA &\textbf{75.3}$\pm$0.96 & \textbf{81.2}$\pm$0.44 & \textbf{91.4}$\pm$0.09 & 69.0$\pm$0.00 & 68.5$\pm$0.17 & \textbf{69.8}$\pm$0.18 & \textbf{73.8}$\pm$0.30 & \textbf{72.5}$\pm$0.19 \\
\hline

\bottomrule
\end{tabular}
\end{table*}

\subsection{Related Work}
\label{section:related_work}
\vspace{-1mm}
Ever since BERT showed its effectiveness on various NLP downstream tasks, there have been more and more methods developed aiming at further improving language pre-training methods' performance on NLP.
Most of them try to achieve this by providing stronger self-supervised signals with existing unsupervised pre-training data~\citep{levine2020pmi,wu2020taking}. Among them, SpanBERT~\citep{joshi2019spanbert} extended BERT by masking contiguous token sequences instead of random tokens during training. The model is trained to predict the masked token sequences correctly.
ELECTRA~\citep{clark2019electra} trains the model to distinguish ``fake" tokens generated a by a generator from the original tokens in the input sequence.
Inspired by ELECTRA, we take a step further. MPA uses the ELECTRA as its backbone, but make more use of the generator's mis-predictions by taking the conflicting context with the mis-predictions into consideration.
Through this way, the mis-predictions act as ill focus detectors and provide language-specific inductive bias as stronger self-supervised information to improve pre-training methods. 

Another line of research aims to improve the efficiency of language pre-training.
\citep{gong2019efficient} observed that parameters in different self-attention layers have similar attention distribution, and proposed a parameter distillation method from shallow layers to deep layers. \citep{xiong2020layer} analyzed Transformer from an optimization perspective and provided a training guideline that significantly reduced pre-training time.
TUPE~\citep{ke2020rethinking} modified how positional embedding is introduced in each self attention layer in BERT, which resulted in both performance and efficiency improvements. Our proposed method is orthogonal to this line of work and can be applied on top of them to further boost their performances.

\end{document}